\documentclass[10pt,twocolumn,letterpaper]{article}

\usepackage[pagenumbers]{cvpr}      
\definecolor{cvprblue}{rgb}{0.21,0.49,0.74}
\usepackage[pagebackref,breaklinks,colorlinks,allcolors=cvprblue]{hyperref}
\usepackage{graphicx}
\usepackage{multicol}
\usepackage{multirow}
\usepackage{makecell}
\newcommand{\com}[1]{\textcolor{red}{#1}}
\usepackage{comment}
\usepackage{wasysym}
\usepackage{algorithm}
\usepackage{algorithmic}
\def\paperID{17396} 
\def\confName{CVPR}
\def\confYear{2026}
\usepackage{graphicx}
\usepackage[font=footnotesize]{caption}

\newcommand{\pipeline}{
\begin{figure*}
  \centering
  \includegraphics[width=1.0\linewidth]{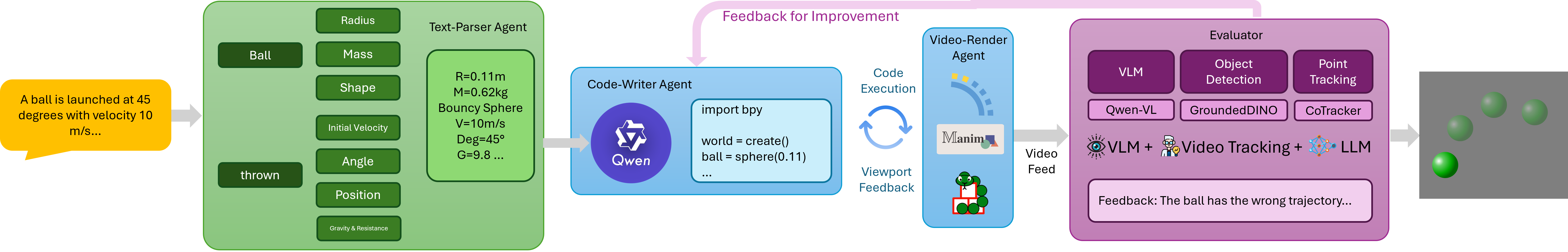}
  \caption{\footnotesize Overview of our multi-agent motion-reasoning engine (MoReGen) for physics-grounded text-to-video synthesis. MoReGen focuses on achieving high-precision Newtonian motion through coordinated multi-agent reasoning. 
Given a natural language prompt, the \emph{text-parser} agent $\mathcal{A}_{\text{text}}$ extracts physical parameters and motion descriptors, which are then translated into executable simulation code by the \emph{code-writer} agent $\mathcal{A}_{\text{coder}}$. 
The resulting code is executed within a sandboxed \emph{video-render} agent $\mathcal{A}_{\text{render}}$ to produce a physically plausible video. This video serves as the basis for evaluator feedback, guiding enhancements in code robustness and physical fidelity for subsequent iterations. By leveraging open-source LLMs, few-shot tuning of $\mathcal{A}_{\text{text}}$ and multi-modal evaluator, MoReGen enables accurate and reproducible Newtonian motion synthesis from natural language instructions.}
  \label{fig:pipeline}
  \vspace{-0.5cm}
\end{figure*}
}

\newcommand{\dataset}{
\begin{figure}
  \centering
  \includegraphics[width=1.0\linewidth]{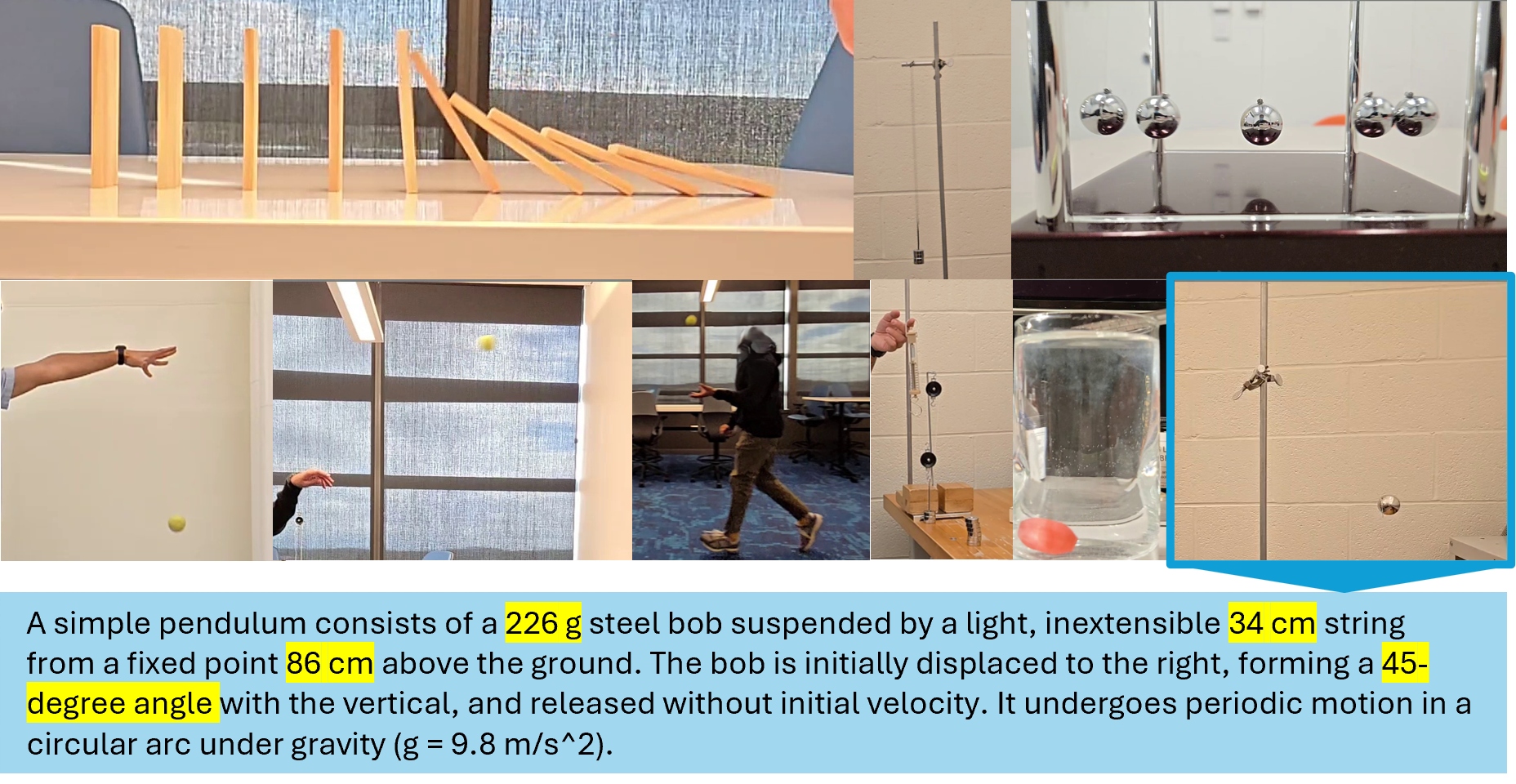}
  \caption{\footnotesize Sample frames and corresponding text prompts from our MoReSet dataset. Each image (extracted from videos) illustrates a distinct Newtonian physics phenomenon. The provided annotation for the pendulum corresponds to the rightmost video in the second row, with highlighted text emphasizing the numerical relationships depicted in the scene.}
  \label{fig:dataset}
   \vspace{-0.5cm}
\end{figure}
}

\newcommand{\quality}{
\begin{figure*}
  \centering
  \includegraphics[width=1.0\linewidth]{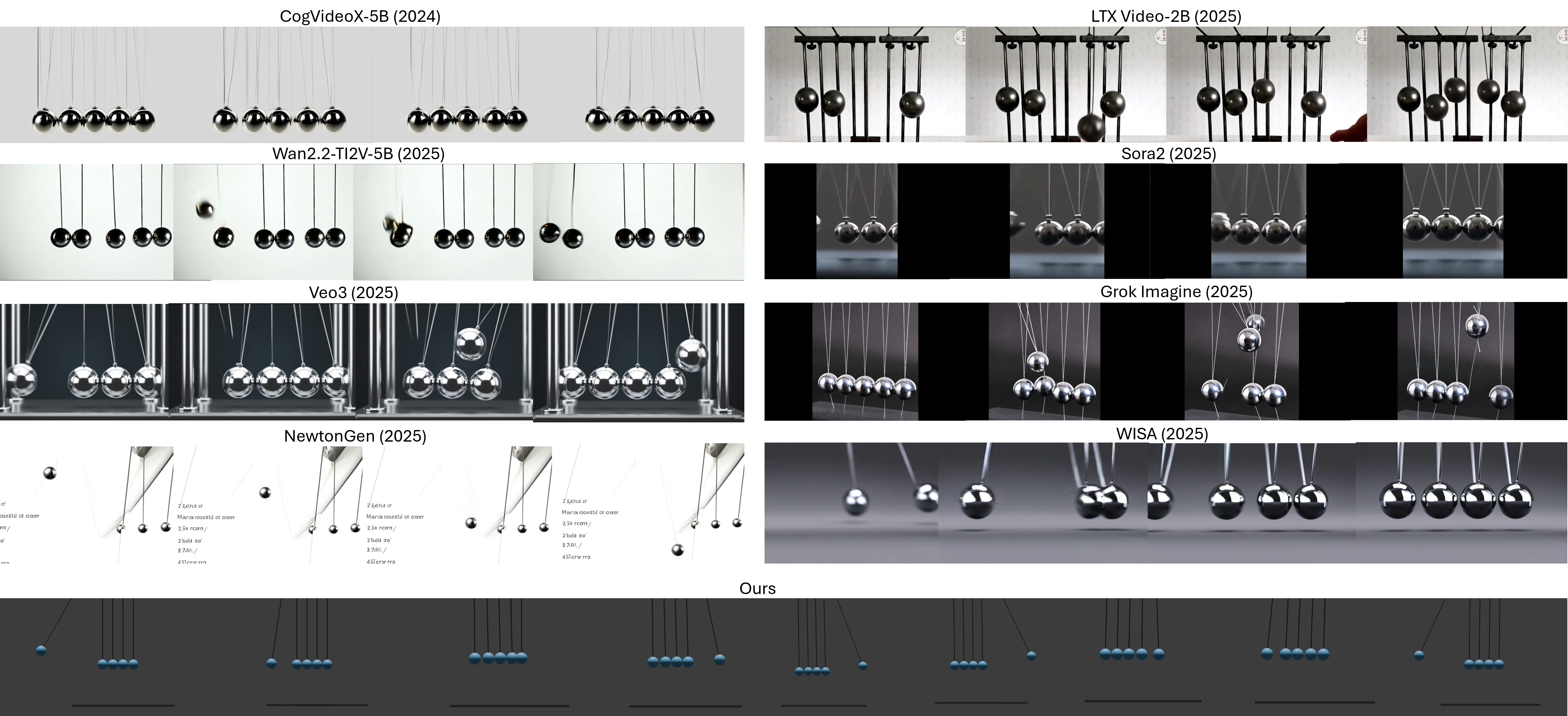}
  \caption{\footnotesize Qualitative comparison of our model with recent open-source and commercial models, prompted to generate a video of Newton's cradle. We used the same prompt across the board: ``Generate a video that showcase the following scene: Five shiny metal balls of a newton's cradle is visible, along with parts of a single vertical string for each metal ball respectively. These strings keeps their respective metal ball suspended. The top part of the newton's cradle is not visible. The camera faces all the five metal balls. The first and leftmost ball is at an angle of 30 degrees from the cradle and released. Due to gravity, the ball comes and strikes the second ball from the left. This causes momentum to be transferred to the fifth and the right most ball which is launched at a slightly lesser angle, having lost some momentum. This process keeps repeating itself till the rightmost ball has lost a lot of momentum when the video ends.''  For Grok Imagine, we always select the first video; for WISA, we use Qwen3-4B to generate asset .json file from our prompt.}
  \label{fig:quality}
\end{figure*}
}

\newcommand{\datasetcomp}{
\begin{table*}[t]
\scriptsize
\caption{Comparison of our dataset with recent physics-based T2V datasets. Symbols indicate whether each dataset contains data for the listed properties: $\checkmark$ denotes inclusion, while $\times$ indicates absence. We define Physics-based Prompts as those that explicitly specify the relevant physical principles and mathematical properties governing the behavior of objects and environments. As of the time of writing, the HQ-Phy dataset has not been publicly released. }
 \vspace{-0.5cm}
\begin{center}
\begin{tabular}{c|ccccccc}
\hline
\makecell{Physics-based\\T2V Datasets} & \makecell{Released\\Year} & \makecell{Video\\Dataset} & \makecell{Video\\Source} & \makecell{Real-world\\Experiment} & \makecell{Physics-based\\Prompts} & \makecell{Object\\Trajectories} & \makecell{Physics\\Phenomenon }\\
\cline{2-8}
\hline
PISABench \cite{li2025pisa}& 2025 & $\checkmark$ & 3D Simulator & $\times$ & $\times$ & $\checkmark$ & 1 \\ 
HQ-Phy \cite{zhang2025think} & 2025 & $\checkmark$ & Web Curated & $\times$ & $\checkmark$ & $\times$ & -\\ 
PhyWorld \cite{kang2024far} & 2024 & $\checkmark$ & 2D simulator & $\times$ & $\times$ & $\checkmark$ & 3 \\
VideoPhy \cite{bansal2024videophy} & 2024 & $\times$ & - & $\times$ & $\times$ & $\times$ & 8\\
T2VPhysBench \cite{guo2025t2vphysbench} & 2025 & $\times$ & - & $\times$ & $\times$ & $\times$ & 12 \\
WISA \cite{wang2025wisa} & 2025 & $\checkmark$ & Web Curated & $\times$ & $\checkmark$ &$\times$ & 17 \\
PhyGenBench \cite{meng2024towards} & 2025 & $\times$ & - & $\times$ & $\times$ & $\times$ & 27 \\
\textbf{MoReGen (Ours)} & 2025 & $\checkmark$ & Manually Collected & $\checkmark$ & $\checkmark$ & $\checkmark$ & 9\\ \hline
\end{tabular}
\label{tbl:dataset_comp}
\end{center}
 \vspace{-0.8cm}
\end{table*}
}

\newcommand{\ourmetric}{
\begin{table}[t]
\scriptsize
\caption{Comparison of object motion trajectories across state-of-the-art T2V models, evaluated using MoRe metrics. To ensure comprehensive object coverage, the initial object detection stage was manually supervised by a human annotator. For objects that did not appear in videos, a random pixel is selected. DTW-N stands for normalized DTW distance, where trajectories are first centered and then scaled down to unit arc length. All videos are downsampled to 480p resolution at 10 frames per second. \textbf{Bold text} indicates best performance.}
 \vspace{-0.5cm}
\begin{center}
\begin{tabular}{c|ccc}
\hline
\multirow{2}{*}{T2V Models} & \multicolumn{3}{c}{\textbf{MoRe Metrics}} \\
\cline{2-4}
& DTW $\downarrow$ & DTW-N $\downarrow$ & Procrustes $\downarrow$ \\
\hline
Wan2.2-TI2V-5B \cite{wan2025wan} & 17.18 $\pm$ 15.88 & 0.10 $\pm$ 0.09 & 0.65 $\pm$ 0.26 \\ 
LTXV-2B-Distilled \cite{hacohen2024ltx} & 16.99 $\pm$ 16.05 & 0.11 $\pm$ 0.11 & 0.71 $\pm$ 0.28 \\ 
CogVideoX-5B \cite{hong2022cogvideo} & 12.94 $\pm$ 13.56 & 0.09 $\pm$ 0.08 & 0.62 $\pm$ 0.26\\
Veo3 \cite{deepmind2024veo} & 13.35 $\pm$ 12.96&0.07 $\pm$ 0.06 & 0.57 $\pm$ 0.27\\
Grok \cite{grok2025imagine} & 13.00 $\pm$ 11.58 & 0.08 $\pm$ 0.08 & 0.55 $\pm$ 0.31\\
Sora2 \cite{brooks2024videoworldsimulators} & 11.21 $\pm$ 12.76 & 0.08 $\pm$ 0.07 & 0.55 $\pm$ 0.30 \\
NewtonGen \cite{yuan2025newtongen} & 17.88 $\pm$ 14.04 & 0.11 $\pm$ 0.09 & 0.62 $\pm$ 0.34\\
WISA \cite{wang2025wisa} & 12.67 $\pm$ 9.22& 0.09 $\pm$ 0.08 & 0.68 $\pm$ 0.27 \\
\textbf{MoReGen (Ours)} &\textbf{8.93} $\pm$ \textbf{9.61} & \textbf{0.06} $\pm$ \textbf{0.07}  & \textbf{0.48} $\pm$ \textbf{0.30}\\ \hline
\end{tabular}
\label{tbl:ourmetric}
\end{center}
 \vspace{-0.8cm}
\end{table}
}

\newcommand{\theirmetric}{
\begin{table}[t]
\scriptsize
\begin{center}
\caption{Comparison of physical validity and consistency across state-of-the-art T2V models (Wan2.2-TI2V-5B \cite{wan2025wan}, LTXV-2B-Distilled \cite{hacohen2024ltx}, CogVideoX-5B \cite{hong2022cogvideo}, Veo3 \cite{deepmind2024veo}, Grok Imagine \cite{grok2025imagine}, Sora2 \cite{brooks2024videoworldsimulators}, NewtonGen \cite{yuan2025newtongen} and WISA \cite{wang2025wisa}) using physics-based video evaluation metrics. For Trajan, videos were resized to 256×256 and 128 points were sampled. AJ denotes average Jaccard, and OA denotes occlusion accuracy. For VideoPhy, each video was rated from 1 to 5 based on semantic adherence (SA) and physics consistency (PC). We report the mean and standard deviation for all evaluated T2V models. \textbf{Bold text} indicates best performance.}
\begin{tabular}{c|cccc}
\hline
 \multirow{2}{*}{T2V Models} & \multicolumn{2}{c}{Trajan \cite{googledeepmind}} & \multicolumn{2}{c}{VideoPhy2 \cite{bansal2025videophy}} \\ \cline{2-5}
& AJ $\uparrow$ & OA $\uparrow$ & SA$\uparrow$ & PC$\uparrow$ \\ \hline
Wan2.2 & 0.38 $\pm$ 0.02 & 0.94 $\pm$ 0.02 & 2.88 $\pm$ 0.46 &  3.77 $\pm$ 0.68\\ 
LTXV & \textbf{0.79} $\pm$ \textbf{0.03} & \textbf{0.99} $\pm$ \textbf{0.02} & 2.95 $\pm$ 0.28 & 3.75 $\pm$ 0.68 \\ 
CogVideoX & 0.17 $\pm$ 0.02 & 0.72 $\pm$ 0.04 & 2.83 $\pm$ 0.47 & 4.07 $\pm$ 0.59\\
Veo3 & 0.15 $\pm$ 0.02 & 0.82 $\pm$ 0.02 & 2.95 $\pm$ 0.33 & 3.49 $\pm$ 0.50\\
Grok & 0.11 $\pm$ 0.02 & 0.72 $\pm$ 0.05 & \textbf{3.04} $\pm$ \textbf{0.34} & 3.76 $\pm$ 0.56 \\
Sora2 & 0.45 $\pm$ 0.02 & 0.92 $\pm$ 0.02 &  3.01 $\pm$ 0.35 & 3.91 $\pm$ 0.52 \\
NewtonGen & 0.56 $\pm$ 0.02 & 0.96 $\pm$ 0.02 & 2.77 $\pm$ 0.42 & 4.21 $\pm$ 0.52\\
WISA &0.41 $\pm$ 0.02& 0.86 $\pm$ 0.03 & 2.92 $\pm$ 0.45 & 4.01 $\pm$ 0.66\\
\textbf{MoReGen (Ours)} & 0.10 $\pm$ 0.03 & 0.64 $\pm$ 0.05 & 2.73 $\pm$ 0.44 & \textbf{4.53} $\pm$ \textbf{0.69}\\\hline
\end{tabular}
\label{tbl:theirmetric}
\end{center}
 \vspace{-0.8cm}
\end{table}
}

\newcommand{\ablation}{
\begin{table*}[t]
\scriptsize
\caption{Comparison of model performance gains from supervised fine-tuning (SFT) and evaluator feedback. Evaluation was conducted using the designated evaluation set from our dataset, with tests performed on Qwen2.5 Coder and GPT-5. Feedback involves one iteration of processing the video to extract object trajectories, comparing them with ground truth, and obtaining an overall evaluation from Qwen2.5-VL. GPT-5 then summarizes this evaluation into a list of actionable improvements, which are fed into the Coder Agent. If the code contains syntax error, GPT-5 is asked to correct the issue without providing the full code. \textbf{Bold text} indicates best performance.}
\vspace{-0.5cm}
\begin{center}
\begin{tabular}{ccc|ccccccc}
\hline
\multicolumn{3}{c}{Ablations} & \multicolumn{3}{c}{MoReGen (Ours)} & \multicolumn{2}{c}{Trajan} & \multicolumn{2}{c}{VideoPhy}\\
\cline{1-10}
Coder Agent & SFT & Feedback & DTW $\downarrow$ & DTW-N $\downarrow$ & Procrustes $\downarrow$ & AJ $\uparrow$ & OA $\uparrow$ & SA $\uparrow$ & PC $\uparrow$\\
\hline
Qwen2.5-Coder-14B & $\times$ & $\times$ &\multicolumn{7}{c}{Unable to generate most cases due to syntax error in provided code.} \\ 
Qwen2.5-Coder-14B & $\times$ & $\checkmark$ & 18.01 $\pm$ 16.53 & 0.08 $\pm$ 0.15 & 0.70 $\pm$ 0.52  & 0.09 $\pm$ 0.02 & 0.61 $\pm$ 0.06 & 2.51 $\pm$ 0.51 & 4.49 $\pm$ 0.57\\
Qwen2.5-Coder-14B &$\checkmark$ & $\times$ &\textbf{8.93} $\pm$ \textbf{9.61} & \textbf{0.06} $\pm$ \textbf{0.07}  & \textbf{0.48} $\pm$ \textbf{0.30} & 0.10 $\pm$ 0.03 & 0.64 $\pm$ 0.05 & \textbf{2.73} $\pm$ \textbf{0.44} & 4.53 $\pm$ 0.69 \\ 
GPT-5 & $\times$ & $\times$ & 15.47 $\pm$ 14.60 & 0.07 $\pm$ 0.07 & 0.58 $\pm$ 0.29 & 0.32 $\pm$ 0.06 & 0.70 $\pm$ 0.04 & 2.60 $\pm$ 0.49 & 4.53 $\pm$ 0.55\\
GPT-5 & $\times$ & $\checkmark$ & 14.13 $\pm$ 14.20& 0.07 $\pm$ 0.07 & 0.51 $\pm$ 0.28 &\textbf{0.38} $\pm$ \textbf{0.06} &\textbf{0.77} $\pm$ \textbf{0.04} &2.52 $\pm$ 0.50 & \textbf{4.57} $\pm$ \textbf{0.52} \\
\hline
\end{tabular}
\label{tbl:ablation}
\end{center}
 \vspace{-0.8cm}
\end{table*}
}

\title{MoReGen: Multi-Agent Motion-Reasoning Engine\\for Code-based Text-to-Video Synthesis}

\author{Xiangyu Bai\textsuperscript{*}\\
\small Northeastern University\\
\small 360 Huntington Avenue, Boston MA 02115\\
{\tt\small bai.xiang@northeastern.edu}
\and
He Liang\textsuperscript{*}\\
\small University of Oxford\\
\small Wellington Square, Oxford OX1 2JD\\
{\tt\small he.liang@cs.ox.ac.uk}
\and
Bishoy Galoaa\\
\small Northeastern University\\
\small 360 Huntington Avenue, Boston MA 02115\\
{\tt\small sobhy.b@northeastern.edu}
\and
Utsav Nandi\\
\small Northeastern University\\
\small 360 Huntington Avenue, Boston MA 02115\\
{\tt\small nandi.u@northeastern.edu}
\and
Shayda Moezzi\\
\small Northeastern University\\
\small 360 Huntington Avenue, Boston MA 02115\\
{\tt\small moezzi.s@northeastern.edu}
\and
Yuhang He \textsuperscript{\dag}\\
\small Microsoft Research\\
\small 725 Granville St., Vancouver, BC, Canada\\
{\tt\small yuhanghe@microsoft.com}
\and
Sarah Ostadabbas \textsuperscript{\dag}\\
\small Northeastern University\\
\small 360 Huntington Avenue, Boston MA 02115\\
{\tt\small s.ostadabbas@northeastern.edu}
}

\begin{document}

\twocolumn[{
\renewcommand\twocolumn[1][]{#1}
\maketitle

\begin{center}
    \vspace{-28pt}
    \includegraphics[width=1.0\linewidth]{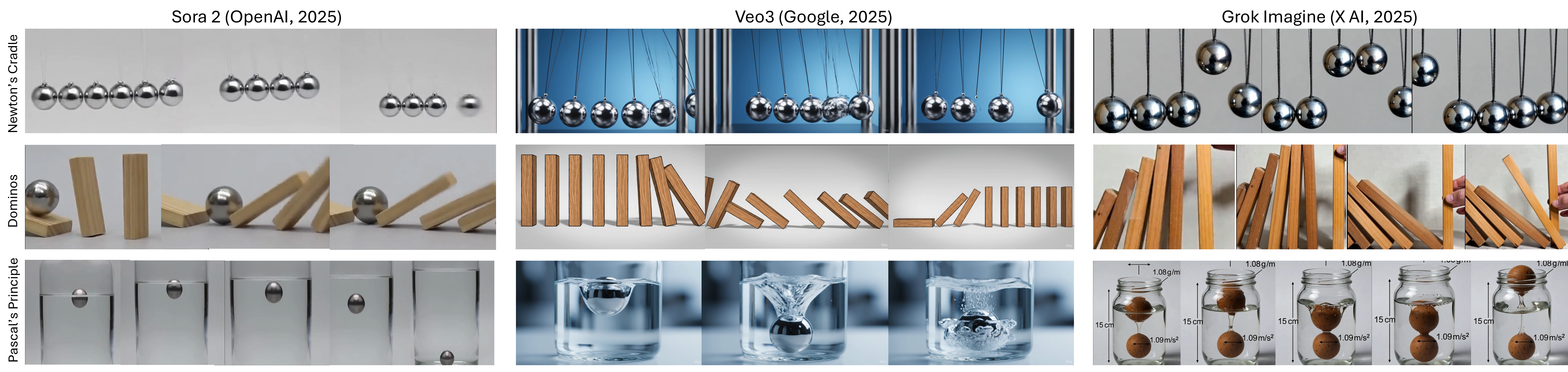}
    \centering
    \captionsetup{type=figure}
    \vspace{-5pt}
   \caption{
    Samples generated from selected prompts in our dataset, MoReSet, using current commercial text-to-video models.
    Left: OpenAI's Sora 2~\cite{brooks2024videoworldsimulators};
    Middle: Google DeepMind's Veo3~\cite{deepmind2024veo};
    Right: X AI's Grok Imagine~\cite{grok2025imagine}.
    The first row illustrates failures in object counting and momentum conservation; the second row highlights misinferred Newtonian forces;  the third shows incorrect velocity and pressure computations.  Even the most advanced text-to-video models lack physical reasoning abilities and are unable to follow physics rules\protect\footnotemark.}
    \label{fig:fail}
\end{center}
}]
\footnotetext[1]{* Equal contribution.}
\footnotetext[2]{\dag Equal contribution corresponding authors.}
\footnotetext[3]{Code, dataset and more samples are available at our GitHub page: \href{https://github.com/ostadabbas/MoReGen-Multi-Agent-Motion-Reasoning-Engine}{https://github.com/ostadabbas/MoReGen-Multi-Agent-Motion-Reasoning-Engine}}

\maketitle
\begin{abstract}
While text-to-video (T2V) generation has achieved remarkable progress in photorealism, generating intent-aligned videos that faithfully obey physics principles remains a core challenge. In this work, we systematically study Newtonian motion-controlled text-to-video generation and evaluation, emphasizing physical precision and motion coherence. We introduce MoReGen, a motion-aware, physics-grounded T2V framework that integrates multi-agent large language models (LLMs), physics simulators, and renderers to generate reproducible, physically accurate videos from text prompts in the code domain. To quantitatively assess physical validity, we propose object-trajectory correspondence as a direct evaluation metric and present MoReSet, a benchmark of 1,275 human-annotated videos spanning nine classes of Newtonian phenomena with scene descriptions, spatiotemporal relations, and ground-truth trajectories. Using MoReSet, we conduct experiments on existing T2V models, evaluating their physical validity through both our MoRe metrics and existing physics-based evaluators. Our results reveal that state-of-the-art models struggle to maintain physical validity, while MoReGen establishes a principled direction toward physically coherent video synthesis.
\end{abstract}

\vspace{-0.1cm}
\section{Introduction}
\vspace{-0.1cm}
\label{sec:intro}
\pipeline
Generative AI has gained unparalleled momentum in recent years, with text-to-video (T2V) models emerging as one of its most intuitive and rapidly advancing applications. These systems translate natural language prompts into rich visual narratives, powering both creative and scientific domains through automated video synthesis \cite{sora,veo,kling,videoldm,wan2025wan}. Despite remarkable progress in visual fidelity, existing T2V approaches remain predominantly appearance-driven rather than physics-grounded. They frequently generate videos that appear realistic yet violate fundamental principles of motion, dynamics, and causality, revealing a critical gap between visual realism and physical plausibility. Addressing this gap requires generative systems that reason about motion and obey physical constraints, not merely replicate static visual patterns (see Figure \ref{fig:fail}).

In this paper, we introduce \textbf{MoReGen}, a multi-agent motion-reasoning engine that redefines text-to-video generation through a motion-aware physics-grounded framework in the code domain. MoReGen unifies multi-agent large language models (LLMs), physics simulators, and rendering engines to generate reproducible and physically accurate videos directly from textual descriptions (see Figure \ref{fig:pipeline}).
In contrast to conventional diffusion-based pipelines that rely on probabilistic denoising, MoReGen directly generates executable code for physics simulators and rendering engines, enabling the synthesis of Newtonian motion dynamics from natural language instructions via a coder LLM agent. To ensure both physical validity and motion coherence, MoReGen employs an iterative multi-agent feedback loop, where motion-understanding models assess temporal consistency and dynamic plausibility, while vision-language models (VLMs) evaluate semantic and visual alignment. 
This interplay among agents fosters context-aware refinement, leading to video generations that are simultaneously physically consistent and visually faithful.

Recent advances in T2V have been largely driven by vision-transformer-based architectures. Researchers have extended pre-trained image diffusion backbones by inflating 2D transformers into 3D modules to integrate  temporal dynamics \cite{bai2025dctdm}. Similar to the ongoing ``arms race'' in LLMs, the dominant paradigm emphasizes architectural scaling through massive datasets and high-cost computational resources \cite{li2025minimax}. While this approach has undeniably improved visual quality, with larger models consistently producing high-fidelity, intent-aligned videos from natural language prompts \cite{dosovitskiy2020image}, their success remains primarily superficial.

Recent evaluation efforts \cite{googledeepmind,guo2025t2vphysbench,meng2024towards} have revealed that, despite impressive aesthetics, most T2V systems still fail to reason about physical reality. Generated videos exhibit spatial and temporal inconsistencies, along with physically impossible motion trajectories \cite{jeong2025track4gen}. The  purely data-driven design of transformer-based architectures dictates that optimization favors pattern memorization over causal reasoning. Instead of inferring the underlying laws that govern motion, these models reproduce statistical correlations in the training data, producing outputs that mirror what was seen rather than reasoning about what should occur \cite{lin2025exploring}. The problem becomes particularly pronounced under out-of-distribution (OOD) prompts, where generated videos deviate sharply from physical reality as Figure \ref{fig:fail} shows. 

A parallel limitation arises in evaluation methodology. Most existing T2V models rely on appearance-based objectives such as Fréchet inception distance (FID) \cite{fid} and Fréchet video distance (FVD) \cite{fvd} to assess performance. While these metrics quantify the distributional similarity of generated pixels to ground-truth data, they provide no indication of whether the generated videos obey underlying physical laws or maintain temporally coherency. As a result, researchers have turned to human-based evaluations \cite{videoldm,bansal2025videophy}; however human perceptual judgments are inherently qualitative and insufficient for evaluating precise physical relationships, limiting their utility in domains where quantitative accuracy is essential \cite{shi2024multi,bastek2024physics,nguyen2024training}. Although recent studies acknowledge this gap, explicit modeling of physical validity remains largely unexplored. Existing physics-informed evaluation methods rely on scene reconstruction metrics such as peak signal-to-noise ratio (PSNR) \cite{liang2024movideo}, which overlook temporal consistency, or on distributional statistics such as semantic adherence \cite{bansal2024videophy} and average Jaccard \cite{googledeepmind} derived from fine-tuned data encoders \cite{meng2024towards}, which are themselves vulnerable to OOD failures.

To address these challenges, MoReGen introduces a paradigm shift toward physics-aware, motion-centric generative modeling. Beyond generating visually plausible videos, MoReGen enforces dynamic coherence, physical validity, and semantic alignment through the interaction of reasoning agents and physics simulators. The framework employs a multi-agent generative architecture in which a text-parser agent converts natural language descriptions into structured physical specifications, a code-writer agent translates those specifications into executable simulation code, a renderer agent generates motion-consistent videos under a physics simulator and a evaluator that analysis generated videos using temporal trajectories, VLMs and LLMs.
The text-parser is supervisedly fine-tuned on phenomenon-specifics datasets to learn robust mappings between linguistic cues and physical parameters. This modular, feedback-driven design enables MoReGen to synthesize dynamically coherent and physically faithful motion sequences that can be quantitatively evaluated.

To quantitatively assess physical validity, we further propose object-trajectory correspondence as a direct evaluation metric. Complementing this, we introduce \textbf{MoReSet}, a benchmark suite compromising 1,275 videos across nine canonical classes of Newtonian phenomena. Each video is human-annotated with descriptions and ground-truth object trajectories. To our knowledge, no existing dataset provides such a detailed pairing of physically grounded annotations with rendered videos. For our experiments, in addition to adopting the physics-based and data-centric metrics proposed by \cite{googledeepmind} and \cite{bansal2025videophy}, we introduce new motion-centric measures, called \textbf{MoRe Metrics} that explicitly quantify trajectory fidelity and physical coherence. Through extensive experiments, we show that MoRe metrics provide a faithful quantitative assessment of physical validity, whereas conventional data-centric metrics fail to capture the motion-level physics accuracy in generated videos. Our key contributions are summarized as follows:
\begin{itemize}
\item We propose MoReGen, a multi-agent text-to-video generation pipeline that directly produces executable code for physics engines, synthesizing physics-grounded videos of Newtonian motion from natural language instructions.
\item We design a modular multi-agent coding and rendering framework that transforms natural language into executable Newtonian simulations, enabling reproducible and physically valid video generation.
\item We introduce MoReSet, a benchmark suite containing 1,275 videos demonstrating nine fundamental Newtonian physics, each paired with detailed scene annotations and ground-truth object motion trajectories.
\item We develop and apply MoRe metrics, a novel motion-centric evaluation metric that reveals a critical research gap between trajectory fidelity and physical coherence in current T2V models, offering a principled foundation for physics-based generative evaluation.
\end{itemize}

\vspace{-0.1cm}
\section{Related Work}
\vspace{-0.1cm}
We review recent research on T2V models and evaluation, emphasizing why data-centric architectures and aesthetics-focused metrics fail to ensure or measure the physical validity. We also highlight the unique advantage of our framework and evaluation over traditional approaches. 
\vspace{-0.1cm}
\subsection{Transformer-based Video Generation}
\vspace{-0.1cm}
Breakthroughs in denoising diffusion probabilistic models (DDPMs) \cite{ho2020denoising} and transformer architectures \cite{attention} have revolutionized video generation. The former enabled realistic synthesis of realistic videos from Gaussian noise \cite{diffusion}, while the latter introduced scalable modeling principles that improve generalization with increasing data and compute \cite{dosovitskiy2020image}. As a result, diffusion-based transformer (DiT) architectures \cite{dit} have come to dominate both industry and academia, setting new standards in open-source generative video modeling \cite{wan2025wan, hong2022cogvideo, hacohen2024ltx, zheng2024open}. Despite their success in visual fidelity, DiT-based models exhibit several fundamental limitations. Their dependence on massive datasets and high-cost computational infrastructure makes large-scale training inaccessible to most researchers, where limited resources pose a significant barrier \cite{peng2025open}. More critically, these models remain appearance-drive rather than physics-aware, they lack an understanding of motion dynamics and physical causality because of their memorization of their training distributions \cite{kang2024far}. When confronted with OOD scenarios, they often produce visually compelling yet physically implausible videos \cite{guo2025t2vphysbench}, revealing a persistent disconnect between aesthetic realism and genuine physical reasoning \cite{bansal2024videophy}. This limitation stems from the inductive biases inherent to transformer architectures, which favor statistical pattern memorization over grounded physical modeling and hinder their generalization beyond the training distribution \cite{xie2025physanimator}. 

\vspace{-0.1cm}
\subsection{Physics-aided Generation and Evaluation}
\vspace{-0.1cm}
Recently, video generation research has begun shifting from purely data-centric and aesthetic alignment approaches towards physics-informed modeling, recognizing that physical laws inherently govern spatial and temporal dynamics in all videos. Bastek et al. \cite{bastek2024physics} proposed a framework that incorporates domain-specific physical guidance into diffusion backbones; Liu et al. \cite{liu2024physgen} used simulators to generate motion based on video priors, and Zhang et al. \cite{zhang2025think} employed LLMs to penalize physically implausible outputs. Lv et al. \cite{lv2024gpt4motion} further advanced this direction by integrating LLM-based planning and simulation to condition diffusion models to enhance motion validity. While these efforts represent meaningful progress toward realism, they remain constrained by the limitations of diffusion backbones, which lack explicit causal reasoning and struggle to maintain full physical consistency. To better evaluate such physics-aware models, several benchmarks and metrics have been introduced. Kang et al. \cite{kang2024far} introduced a 2D simulation testbed for assessing object movement and collisions; Bansal et al. \cite{bansal2024videophy} proposed VideoPhy, a benchmark evaluating physical validity through physical commonsense reasoning; and Guo et al. \cite{guo2025t2vphysbench} presented T2VPhysBench, where human evaluators assess model adherence to twelve core physical laws. Although valuable, these approaches often rely on LLM- or VLM-based captioning and scoring pipelines, which introduce risks of hallucination, bias, and lack of grounded understanding, limiting their quantitative reliability.

 In this paper, we address these challenges through MoReGen, a physics-grounded video generation framework that integrates reasoning agents with simulation engines to ensure physically coherent synthesis. Complementing it, we introduce MoReSet, a benchmark capturing a broad spectrum of Newtonian phenomena with dense annotations of motion trajectories and spatiotemporal relations. Together with MoRe Metrics, a suite of motion-centric evaluation metrics grounded in physical trajectory, our framework establishes a principled approach to both generating and evaluating physics-valid T2V models.


\vspace{-0.1cm}
\section{Introducing MoReGen}
\vspace{-0.1cm}
\label{sec:macode}

The MoReGen framework is designed to achieve physics-precise T2V generation by explicitly modeling Newtonian motion through multi-agent collaboration. Given a natural language prompt, MoReGen produces executable simulation code that accurately represents the Newtonian dynamics. MoReGen consists of three cooperative agents and an evaluator: a \emph{text-parser} agent $\mathcal{A}_{\text{text}}$, a \emph{code-writer} agent $\mathcal{A}_{\text{coder}}$, a \emph{video-render} agent $\mathcal{A}_{\text{render}}$ and multi-component evaluator $\mathcal{E}$ ~(see Figure ~\ref{fig:pipeline}). All three agents are LLM-based (Qwen in our case). As illustrated in Algorithm~\ref{alg:more},
$\mathcal{A}_{\text{text}}$ parses the raw description into a structured Newtonian specification, and $\mathcal{A}_{\text{coder}}$ converts this specification into executable physics simulation code, which is run inside a sandboxed environment to obtain object configurations and trajectories.
$\mathcal{A}_{\text{render}}$ then produces a rendering script that consumes these trajectories to generate the video and the evaluator iteratively refines the video by providing feedback to $\mathcal{A}_{\text{coder}}$. By leveraging the reasoning and code-generation capabilities of LLMs, MoReGen can generate high-fidelity Newtonian motion videos that are physically consistent, requiring only few-shot training $\mathcal{A}_{\text{text}}$.


\vspace{-0.1cm}
\subsection{Supervised Fine-Tuned Text-Parser Agent}
\vspace{-0.1cm}
Given a free-form textual description of a Newtonian phenomenon $x$ (e.g., “a ball is launched at 45$^\circ$ with velocity 10\,m/s”), the text-parser agent $\mathcal{A}_{\text{text}}$ (green block in Figure~\ref{fig:pipeline}) produces a structured simulation specification containing all required objects, parameters, and initial conditions. This task is challenging because (1) natural-language descriptions are often incomplete with missing parameters and conditions that are not explicitly stated, and (2) different physical processes rely on distinct governing variables--such as ramp inclination for slope, and launch velocities for projectiles. Without explicit supervision, generic prompts rarely specify sufficient constraints, leading untrained LLMs to merely infer physical parameters and generate incomplete or inconsistent representations.

To provide precise supervision for training, we first construct a phenomenon-specific structured schema for each class of Newtonian motion. These schemas serve as the target output format during supervised fine tuning (SFT) and specify (1) valid parameter ranges and physical units, (2) the essential physical entities (e.g., objects, anchors, constraints), (3) geometric and mechanical consistency requirements, and (4) physically coherent initial conditions. Each schema is instantiated through a task-specific prompt, which reliably produces high-quality structured specifications from raw textual descriptions. All generated outputs are manually validated to ensure physical correctness, parameter consistency, and conformity to the schema; the resulting (text, specification) pairs form the SFT training data.

We adopt SFT to enable $\mathcal{A}_{\text{text}}$ to internalize the mapping from natural language to structured physics specifications, instead of relying on the task-specific prompts or schema templates. After fine-tuning, the agent no longer requires phenomenon-specific prompts: a single general instruction suffices to generate complete and physically consistent specifications across all phenomena. During training, the model learns lightweight linguistic reasoning rules (e.g., interpreting ``first two balls” or ``pushed from the right” into object indices and force directions) and acquires the ability to infer missing parameters while avoiding hallucinated values. We fine-tune Qwen2.5-Coder-14B~\cite{qwen2025qwen25technicalreport} on 1200 curated text–specification pairs using AdamW (learning rate $1\times 10^{-5}$) for five epochs. The model is optimized with supervised next-token loss:
\begin{equation}
\mathcal{L}_{\text{SFT}}
= -\, \mathbb{E}_{(x, S)\sim\mathcal{D}}
\left[
    \sum_{t=1}^{|S|}
        \log p_\theta \!\left(S_t \mid x, S_{<t}\right)
\right],
\end{equation}
where $S$ is the target structured specification paired with input  description $x$ sampled from all training pairs $\mathcal{D}$.

The fine-tuned model thus generates complete and physically coherent structured specifications, even under under-specified or diversely phrased descriptions. These structured outputs then serve as the authoritative input for subsequent code synthesis and video rendering.

\vspace{-0.1cm}
\subsection{Code-Writer and Video-Render Agents}
\vspace{-0.1cm}
The middle stage of our pipeline (blue block in Figure~\ref{fig:pipeline}) is responsible for converting the structured specification into an executable physics simulation and a rendered video. This stage is implemented by the code-writer agent $\mathcal{A}_{\text{coder}}$, which synthesizes the simulation code, and the video-render agent $\mathcal{A}_{\text{render}}$, which generates the rendering script and produces the final video from simulated trajectories.

\noindent \textbf{Code-Writer Agent ($\mathcal{A}_{\text{coder}}$).} The code-writer agent receives the structured specification from $\mathcal{A}_{\text{text}}$ as input and generates executable simulation code $C_t$ that simulates the underlying Newtonian motion process. The code is based on a unified prompt that provides a general Python class framework defining the key simulation components, including space initialization, object creation, constraint setup, and simulation loop. Within this framework, the agent reads each field from the structured specification, assigns corresponding numerical parameters, and selects appropriate open-source physics engine APIs to instantiate bodies, shapes, and constraints. The generated code saves all object configurations and records the system states at every simulation step $\Delta t$, including position, velocity, and orientation, forming a complete temporal trajectory for each object. By leveraging such physics engines ~(\textit{e.g.} Pymunk~\cite{pymunk}, Blender~\cite{blender} or Manim~\cite{manim}), the generated code models Newtonian motion physics precisely, ensuring that the resulting simulation adheres to real-world mechanics rather than statistical approximations.

\noindent \textbf{Video-Render Agent ($\mathcal{A}_{\text{render}}$).} The rendering agent executes the code $C_t$ within a sandboxed environment and produces two synchronized outputs: (1) \emph{telemetry data} recording each asset's spatial position, velocity, and orientation over time; (2) \emph{rendered video} $v_t$ that visually depicts the simulated motion. The rendering stage is jointly synthesized by $\mathcal{A}_{\text{render}}$ through a single general-purpose prompt, requiring no task-specific adaptation. Since visualization essentially involves reconstructing simulated geometries and replaying state trajectories, the same rendering logic generalizes naturally across all physical phenomena. The generated module employs \texttt{pygame}—which integrates seamlessly with Pymunk—to interpret the simulation configuration, rebuild scene geometry, and replay the recorded object states in real time. This unified rendering pipeline produces videos $v_t$ that not only reproduce the simulated Newtonian motion with high temporal and spatial fidelity but also ensure visual coherence across diverse motion types. Besides, this decoupled design between trajectory generation and video rendering allows the framework to be easily extended to more realistic 3D rendering engines such as Unreal Engine, Blender, or Unity, enabling future integration with photorealistic simulation environments.

\begin{algorithm}[t]
\caption{MoReGen Pipeline}
\label{alg:more}
\begin{algorithmic}[1]
\REQUIRE Raw text prompt $x$
\ENSURE Rendered video $v_t$ and telemetry $\mathcal{T}$

\STATE \textbf{Text Parsing}
\STATE \quad $\mathcal{S} \gets \mathcal{A}_{\text{text}}(x)$ // produce structured specification

\STATE \textbf{Code Synthesis}
\STATE \quad $C_t \gets \mathcal{A}_{\text{coder}}(\mathcal{S})$ // generate physics simulation code

\STATE \textbf{Physics Simulation in Sandbox}
\STATE /* get scene $\mathcal{C}$ and trajectories $\mathcal{T}$ from simulation */
\STATE \quad $(\mathcal{C}, \mathcal{T}) \gets \mathrm{Simulate}(C_t)$ 

\STATE \textbf{Video Rendering}
\STATE \quad $v_t \gets \mathcal{A}_{\text{render}}(\mathcal{C}, \mathcal{T})$ //render video from simulation

\RETURN $v_t, \mathcal{T}$
\end{algorithmic}
\label{alg:pipeline}
\vspace{-0.1cm}
\end{algorithm}

\vspace{-0.1cm}
\subsection{Evaluator Design}
\vspace{-0.1cm}
We introduce a multi-iteration refinement process for MoReGen to enhance the robustness and physical fidelity of the generated code and video. The evaluator (colored purple in Figure \ref{fig:pipeline}) analysis output videos ($v_t$) and provide feedback to $\mathcal{A}_{\text{coder}}$ to guide the refinement process. First, we detect object locations in the video using GroundedDINO \cite{liu2024grounding} informed by object descriptors extracted from $\mathcal{A}_{\text{text}}$. These locations guide CoTracker3 \cite{karaev2025cotracker3} to estimate normalized object trajectories $\mathcal{T}_{est}$, after which an LLM (Qwen in this case) evaluates the similarity and divergence from trajectory $\mathcal{T}$ obtained from simulation process. We also use Qwen2.5-VL \cite{bai2025qwen2} to assesses the video from two perspectives: physical plausibility based on a set of physics rules, and alignment with the intended prompt $x$. Finally, we use LLM to synthesize feedback $\mathcal{F}$ by summarizing trajectory alignment, physical correctness, and prompt fidelity:
\begin{equation}
\begin{split}
    \mathcal{F}=LLM(\text{``Summarize"}, <LLM(\text{``Traj''}, \mathcal{T}_{est}, \mathcal{T}), \\VLM(\text{``Phys''}, v_t), VLM(\text{``Intent''}, v_t, x)>).
\end{split}
\end{equation}

This data is provided to $\mathcal{A}_{\text{coder}}$ to guide improvements in both code generation and video synthesis for next iteration.
 \vspace{-0.1cm}
\section{Experimental Design and Results}
 \vspace{-0.1cm}
\label{sec:experiment}
This section presents the MoReSet benchmark and MoRe metrics, a trajectory-centric evaluation framework quantifying the physical validity of generated videos beyond visual realism. We demonstrate that existing benchmarks and metrics fail to capture fundamental Newtonian principles and show that MoReGen consistently preserves both user intent and underlying physical relationships. Finally, we conduct ablation studies to assess the contribution of fine-tuning, model scale, and evaluator feedback to the physics accuracy of generated videos.

 \vspace{-0.1cm}
\subsection{MoReSet Benchmark}
 \vspace{-0.1cm}

We introduce MoReSet, a novel open-source benchmark for evaluating T2V models through the lens of physics fidelity. The dataset comprises of 1,275 videos spanning nine fundamental categories of Newtonian physics experiments: gravity, acceleration, collision, oscillation, momentum, buoyancy, inertia, pendular motion and pulley mechanics. The training set includes 1,200 Blender-generated simulations each paired with a free-form textual description and a high-quality structured JSON specification  $\mathcal{S}$ capturing complete physical parameters. The test set includes 75 real-world laboratory videos, each human-annotated with: a natural-language description detailing the scene composition, physical relationships between objects and their environment, and chronological dynamics; object-level labels and environmental attributes; and the camera perspective.
To enable precise motion-based evaluation, every test video is annotated with key object identities and full pixel-level trajectories, extracted automatically via CoTracker3 \cite{karaev2025cotracker3}, supervised and corrected by human annotators. Figure \ref{fig:dataset} shows representative samples, and Table \ref{tbl:dataset_comp} compares MoReSet with existing benchmarks for physics-enabled T2V model evaluations.  Unlike prior datasets that target limited physics scenarios or rely on synthetic motion heuristics, MoReSet offers broad physical coverage and explicit trajectory annotations, providing  a standardized testbed for quantitative, physics-grounded T2V evaluation.


\dataset
\datasetcomp
 \vspace{-0.1cm}
\subsection{Evaluation Metrics}
 \vspace{-0.1cm}
 Existing physics-aware metrics remain inadequate for assessing  physical validity, as they emphasize perceptual similarity or in-distribution priors over causal motion accuracy. 

\noindent \textbf{MoRe Metrics.} Instead of pixel-level similarity metrics (FID, FVD, PSNR, and LPIPS) or latent-space features from pretrained video encoders (Trajan, VideoPhy), we propose MoRe metrics, a trajectory-based suite comprised of dynamic time warping (DTW), Normalized DTW (DTW-N) and Procrustes Analysis. These metrics directly measure the trajectory fidelity of key objects in motion.
Given a structured specification $\mathcal{S}$, we identify the key object in motion, localize it in the initial frame via text-grounded object detection, and tracks its trajectory throughout the video. We then compute: (1) DTW \cite{senin2008dynamic} to align estimated trajectories with ground-truth sequences of varying temporal lengths by projecting them into a non-linear space; (2) DTW-N to normalize distance across clips of different scales and durations; and (3) Procrustes Analysis Score  \cite{gower1975generalized} to assess geometric and structural alignment between predicted and ground-truth trajectories. 


\noindent \textbf{Data-Driven Physics Metrics} To maintain comparability with community baselines, we also report results using two state-of-the-art physics-based evaluation frameworks from Trajan \cite{googledeepmind} and VideoPhy2 \cite{bansal2024videophy}. Trajan employs a point-track autoencoder trained to reconstruct spatiotemporal trajectories from BootsTAP \cite{doersch2024bootstap}, reporting average Jaccard (AJ) and Occlusion accuracy (OA) to measure temporal consistency and object visibility, respectively. VideoPhy2 provides Semantic Adherence (SA) and Physical Commonsense (PC) scores via its AutoEvaluator, a pretrained video-language model tuned on VidCon \cite{Bansal_2024_CVPR} for physics understanding. 

Together, these methods allow us to evaluate not only visual fidelity and semantic alignment but also the underlying physical coherence of generated sequences.

\vspace{-0.1cm}
\subsection{Evaluation with MoRe Metrics}
\vspace{-0.1cm}
To capture the breadth of model capabilities, we evaluate three categories of T2V models using MoRe metrics on MoReSet test prompts: (1) medium-sized open-source models (Wan2.2 \cite{wan2025wan}, LTX Video \cite{hacohen2024ltx} and CogVideoX \cite{hong2022cogvideo}), representing academically driven, general-purpose T2V approaches; (2) large-scale commercial models (Veo3 \cite{deepmind2024veo}, Sora2 \cite{brooks2024videoworldsimulators} and Grok Imagine \cite{grok2025imagine}), reflecting the current state-of-the-art; and (3) physics-based models that support text-only input (NewtonGen \cite{yuan2025newtongen} and WISA \cite{wang2025wisa}), used as a baseline for comparison with MoReGen to expose their limitations and underscore our superior physics consistency. All models are evaluated using identical prompts from the MoReSet test set. As seen in Table \ref{tbl:ourmetric}, MoReGen outperforms existing state-of-the-art T2V engines by a large margin, indicating better alignment with ground truth object trajectories. These results align with qualitative analyses and human observations, enabled by the integration of a physics engine that produces realistic, physically accurate object motion. Additionally, we observe that while commercial T2V engines significantly outperform smaller models in visual aesthetics and overall quality, this advantage does not extend to motion trajectory accuracy. Notably, CogVideoX-5B and WISA, both of which are medium-sized open-source model, achieves comparable performance to OpenAI’s Sora2 \cite{brooks2024videoworldsimulators} when evaluated using MoRe metrics. Similarly, NewtonGen and Wan2.2 exhibit comparable accuracy--an expected outcome given that NewtonGen and WISA are derived from Wan and CogVideo. This suggests that their approach to fine-tuning existing T2V models for physics validity is ineffective. Another notable observation is the poor stability of existing models when prompted with varying physics principles, as evidenced by large standard deviations in their performance. This variability suggests an imbalance in their training data, with different models excelling in different physical domains. In contrast, MoReGen maintains consistent performance across principles, owing to its rule-driven design and physics-grounded generation pipeline.
\ourmetric

 \vspace{-0.1cm}
\subsection{Evaluation with Data-Driven Physics Metrics}
 \vspace{-0.1cm}
We further evaluate MoReGen and competing models on the MoreSet test set using data-driven physics metrics adopted from Trajan and VideoPhy2, to reveal their limitations. As shown in table \ref{tbl:theirmetric}, both frameworks fail to provide reliable assessments under OOD conditions, consistently assigning lower scores to our simulator-rendered videos due to their divergence from the metrics' training distributions.
Trajan fails to capture either motion correctness or perceptual realism as it awards disproportionately high scores to low-fidelity models such as LTX video that struggle to generate objects that even loosely resemble the user prompt, let alone produce accurate motion. The performance of LTX video according to this metric even surpasses larger commercial models (Veo3, Grok Imagine) that exhibit clearer physical plausibility. 
For VideoPhy2, the Semantic Adherence (SA) metric strongly correlates with visual quality rather than physical validity, ranking visually appealing but physically inconsistent models, namely Sora 2 and Grok Imagine, highest. 
While the Physical Commonsense (PC) metric better reflects actual dynamics, correctly favoring MoreGen and other simulator-based models, it still misranks CogVideoX above more physically accurate systems. In our evaluation, we observed that both Sora 2 and Grok Imagine produce more realistic physical motions than CogVideoX, however the PC metric fails to capture this. These observations confirm that current ``physics-based'' metrics fail to robustly capture the physical validity of generated videos.

\theirmetric

 \vspace{-0.1cm}
\subsection{Qualitative Analysis}
 \vspace{-0.1cm}
\quality
In Figure \ref{fig:quality}, we compare our generation against state-of-the-art open-source and commercial T2V models using Newton's cradle, which illustrates collision and momentum. Qualitatively, the following observations can be made: first, most data-driven T2V models struggle with object counting, often generating an incorrect number of balls. When requested with five balls, LTX Video \cite{hacohen2024ltx}, Sora2 \cite{brooks2024videoworldsimulators}, Veo3 \cite{deepmind2024veo}, NewtonGen \cite{yuan2025newtongen} and WISA \cite{wang2025wisa} all created four balls in the beginning, and while CogVideoX \cite{hong2022cogvideo} and Wan2.2 \cite{wan2025wan} created the corrected number of balls, Wan2.2 introduced a sixth ball mid-sequence, and CogVideoX produced a largely static video. Secondly, many models misinterpret the collision dynamics and momentum transfer as object addition or removal. This is particularly evident in LTX Video, Wan2.2, Veo3, and Grok Imagine \cite{grok2025imagine}, where a ball is inserted into the cradle perpendicular to the intended motion axis (left-right). Lastly, although NewtonGen and WISA failed to capture the correct number of balls and concept of Newton's cradle, their physics-optimized design enables motion trajectories that appear physically plausible to the human eye; other models consistently introduced motion artifacts and physically inconsistent behaviors in their video outputs: In Wan2.2, the ball merges with the others and then splits again; in Sora, the system halts after the first ball strike; and in Veo3, the ball is lifted up while the string leans unnaturally to the right. In contrast, videos produced by our model demonstrate superior physical understanding, yielding a realistic Newton's cradle with the correct number of balls and physics-valid motion. Notably, it captures subtle oscillations of the middle balls due to imperfect impact--details typically observed only in real-world scenarios.


 \vspace{-0.1cm}
\subsection{Ablation Study}
 \vspace{-0.1cm}
 \ablation
To assess the contributions of individual model components, we conduct experiments across varying configurations of the Coder Agent and the presence or absence of supervised fine-tuning (SFT) and evaluator feedback (see Table \ref{tbl:ablation}). We compare the performance of a medium-sized coding model, Qwen2.5-Coder-14B, against GPT-5, and demonstrate that incorporating SFT significantly enhances the physical accuracy of generated videos--enabling the 14B model to outperform GPT-5 according to our evaluation metrics. We also found that incorporating even a single iteration of evaluator feedback significantly improves model performance. Initially, the 14B variant was unable to generate code executable by the rendering client. However, after adding feedback, Qwen successfully produced runnable code. Feedback also led to consistent performance gains for GPT-5 across all evaluation metrics. Finally, consistent with our observations in Section 4.4, Trajan and VideoPhy2 show disagreement with our metrics on the physics validity of the videos. However, physics consistency (PC) remains strong across all code-driven entries, consistently outperforming data-driven T2V models. Despite the accuracy of the physics engine, semantic adherence (SA), average Jaccard (AJ), and occlusion accuracy (OA) remain consistently low across all models, highlighting a bias in these metrics toward data- and aesthetics-driven approaches.


 \vspace{-0.1cm}
\section{Conclusion}
 \vspace{-0.2cm}
In this paper, we introduced MoReGen, a motion-aware, physics-grounded text-to-video (T2V) framework that unifies multi-agent LLMs, simulators, and renderers to generate reproducible and physically accurate videos from text prompts in the code domain. Additionally, we proposed MoRe Metrics as a direct evaluation approach and presented MoReSet, a benchmark comprising human-annotated videos spanning nine classes of Newtonian phenomena with scene descriptions, spatiotemporal relations, and ground-truth trajectories for quantitative assessment of physics validity. Finally, we demonstrated that state-of-the-art models struggle with physical consistency, while our framework provides a principled path toward physically coherent video synthesis. MoReGen marks a new direction for physics-grounded T2V research, and we plan to extend it toward photorealistic generation in future work. 

{
    \small
    \bibliographystyle{ieeenat_fullname}
    \bibliography{main}
}

\end{document}